\documentclass[sigconf]{acmart}
\usepackage{bm}
\usepackage{bbm}
\usepackage{makecell}
\usepackage{colortbl}

\AtBeginDocument{%
  \providecommand\BibTeX{{%
    \normalfont B\kern-0.5em{\scshape i\kern-0.25em b}\kern-0.8em\TeX}}}

\setcopyright{acmcopyright}
\copyrightyear{2021}
\acmYear{2021}
\acmDOI{10.1145/3474085.3475447}

\acmConference[MM '21] {Proceedings of the 29th ACM Int'l Conference on Multimedia}{October 20--24, 2021}{Chengdu, Sichuan Province, China.}
\acmBooktitle{Proceedings of the 29th ACM Int'l Conference on Multimedia (MM '21), Oct. 20--24, 2021, Chengdu, Sichuan Province, China}
\acmPrice{15.00}
\acmISBN{978-1-4503-8651-7/21/10}

\acmSubmissionID{1555}

\begin{document}
\fancyhead{}

\title{InsPose: Instance-Aware Networks for Single-Stage Multi-Person Pose Estimation}


\author{Dahu Shi}
\affiliation{%
  \institution{Hikvision Research Institute}
  \city{Hangzhou}
  \country{China}
}
\email{shidahu@hikvision.com}

\author{Xing Wei}
\affiliation{%
  \institution{School of Software Engineering}
  \institution{Xi'an Jiaotong University}
}
\email{weixing@mail.xjtu.edu.cn}

\author{Xiaodong Yu}
\affiliation{%
  \institution{Hikvision Research Institute}
  \city{Hangzhou}
  \country{China}
}
\email{yuxiaodong7@hikvision.com}

\author{Wenming Tan}
\affiliation{%
  \institution{Hikvision Research Institute}
  \city{Hangzhou}
  \country{China}
}
\email{tanwenming@hikvision.com}

\author{Ye Ren}
\affiliation{%
  \institution{Hikvision Research Institute}
  \city{Hangzhou}
  \country{China}
}
\email{renye@hikvision.com}

\author{Shiliang Pu}
\authornote{Corresponding author.}
\affiliation{%
  \institution{Hikvision Research Institute}
  \city{Hangzhou}
  \country{China}
}
\email{pushiliang.hri@hikvision.com}

\begin{abstract}
Multi-person pose estimation is an attractive and challenging task.
Existing methods are mostly based on two-stage frameworks, which include top-down and bottom-up methods.
Two-stage methods either suffer from high computational redundancy for additional person detectors or they need to group keypoints heuristically after predicting all the instance-agnostic keypoints.
The single-stage paradigm aims to simplify the multi-person pose estimation pipeline and receives a lot of attention.
However, recent single-stage methods have the limitation of low performance due to the difficulty of regressing various full-body poses from a single feature vector.
Different from previous solutions that involve complex heuristic designs, we present a simple yet effective solution by employing instance-aware dynamic networks.
Specifically, we propose an instance-aware module to adaptively adjust (part of) the network parameters for each instance.
Our solution can significantly increase the capacity and adaptive-ability of the network for recognizing various poses, while maintaining a compact end-to-end trainable pipeline.
Extensive experiments on the MS-COCO dataset demonstrate that our method achieves significant improvement over existing single-stage methods, and makes a better balance of accuracy and efficiency compared to the state-of-the-art two-stage approaches.
The code and models are available at \url{https://github.com/hikvision-research/opera}.
\end{abstract}




\maketitle

\begin{figure}[ht]
  \centering
  \includegraphics[width=\linewidth]{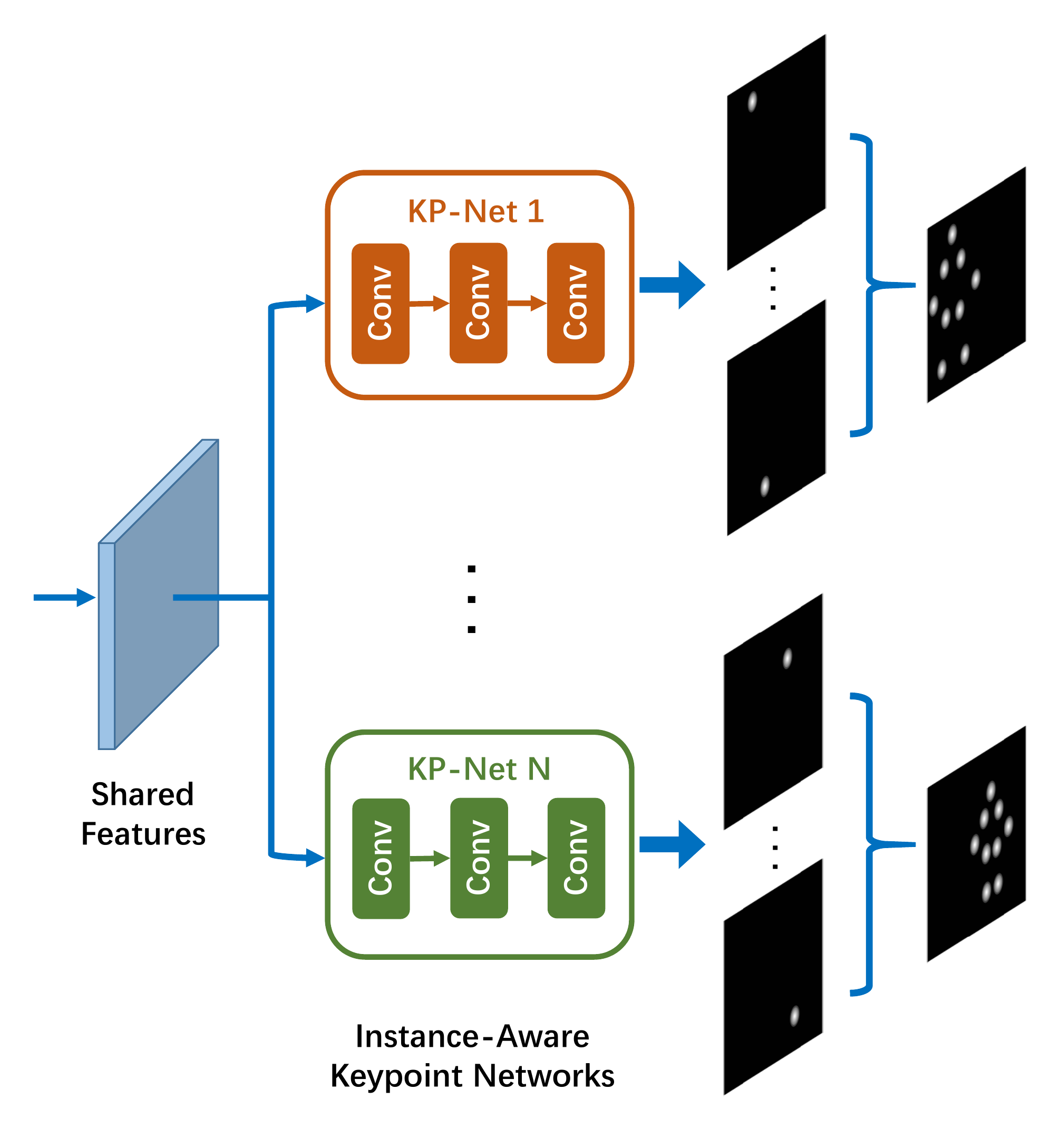}
  \vspace{-0.8cm}
  \caption{InsPose makes use of instance-aware keypoint networks (KP-Nets) to directly predict the full-body pose for each instance.
  }
  \label{fig:CondPosePerInstance}
\end{figure}

\section{Introduction}
Multi-person pose estimation aims to identify all the instances and detect the keypoints of each person simultaneously.
It is a fundamental task and has a wide range of applications such as motion recognition \cite{wang2018rgb, wang2013approach}, person re-identification \cite{li2018harmonious}, pedestrian tracking \cite{zhu2013apt}, and athletic training assistance~\cite{DBLP:conf/mm/WangQPFZ19}.

Multi-person pose estimation methods are usually based on two-stage frameworks including the bottom-up and top-down approaches.
Bottom-up methods \cite{cao2017realtime, pishchulin2016deepcut, insafutdinov2016deepercut, newell2016associative, papandreou2018personlab} first detect all the potential keypoints in the image in an instance-agnostic manner, and then perform a grouping process to get the final instance-aware poses.
Usually, the grouping process is heuristic in which many tricks are involved and does not facilitate end-to-end learning.
On the other hand, top-down methods \cite{he2017mask, fang2017rmpe, chen2018cascaded, xiao2018simple, sun2019deep, dai2021rsgnet, DBLP:conf/mm/JiangHZWX20, DBLP:conf/mm/BaoLHD020} first detect each individual by a bounding-box and then transfer the task to an easier single-person pose estimation problem. 
While avoiding the heuristic grouping process, top-down methods come with the price of high computational complexity as mentioned in ~\cite{nie2019single}.
Especially, the running time of top-down methods depends on the number of instances contained in the image.
Similarly, top-down methods are also not learned in an end-to-end fashion.

Recently, the single-stage pose estimation paradigm~\cite{nie2019single, tian2019directpose, wei2020point} is proposed and receives a lot of attention.
This single-stage framework receives great interest since 1) it directly learns a mapping from the input image to the desired multi-person poses and is end-to-end trainable, and 2) it can consolidate object detection and keypoint localization in a simple and unified framework, and thus 3) presents a more compact pipeline and attractive simplicity and efficiency over two-stage methods.

A straightforward way for single-stage pose estimation is adapting the bounding-box regression head of the single-stage object detectors~\cite{tian2019fcos, zhang2020bridging}.
However, such a naive approach expects a single feature vector (namely, a point feature) to regress the precise locations of all the keypoints.
While a point feature may be sufficient for bounding box regression to some extent, it is difficult to encode rich pose information considering that keypoints have much more geometric variations than the regular box corners.
As a result, the pose estimation accuracy of this approach is unsatisfactory~\cite{tian2019directpose}.
To deal with this problem, SPM~\cite{nie2019single} proposes a progressive joint representation by dividing body joints into hierarchies induced from articulated kinematics.
CenterNet~\cite{zhou2019objects} proposes a hybrid solution by first regressing joint locations and then refining with the help of additional heatmaps.
DirectPose~\cite{tian2019directpose} proposes a keypoint alignment module with the aim of aligning the point feature to the features around target joints.
However, they all involve heuristic designs and still result in less accurate performance compared to the state-of-the-art bottom-up method~\cite{cheng2020higherhrnet}, leaving room for further optimization.

In this paper, we present a direct and simple framework for single-stage multi-person pose estimation, termed \emph{InsPose}.
The core of our method is an instance-aware module to adaptively adjust (part of) the network parameters for each instance, as illustrated in Figure~\ref{fig:CondPosePerInstance}.
Instead of using a standard keypoint network (called, KP-Nets) with fixed parameters for predicting all instances of various poses, we borrow the idea from recent dynamic neural networks~\cite{yang2019condconv,Chen_2020_CVPR,DBLP:conf/nips/DenilSDRF13,jia2016dynamic} to generate instance-aware parameters.
Specifically, these KP-Nets are dynamically generated with respect to the different spatial locations of the input.
Afterwards, they are applied to the whole feature maps to directly predict the full-body poses, get rid of the need for additional handcrafted processing.
KP-Nets can better encode all kinds of pose variants since each one of them only focuses on an individual instance.
Moreover, each KP-Net only consists of three $1\times 1$ convolutions, leading to a compact and efficient architecture.

\begin{figure*}[ht]
  \centering
  \includegraphics[width=\textwidth]{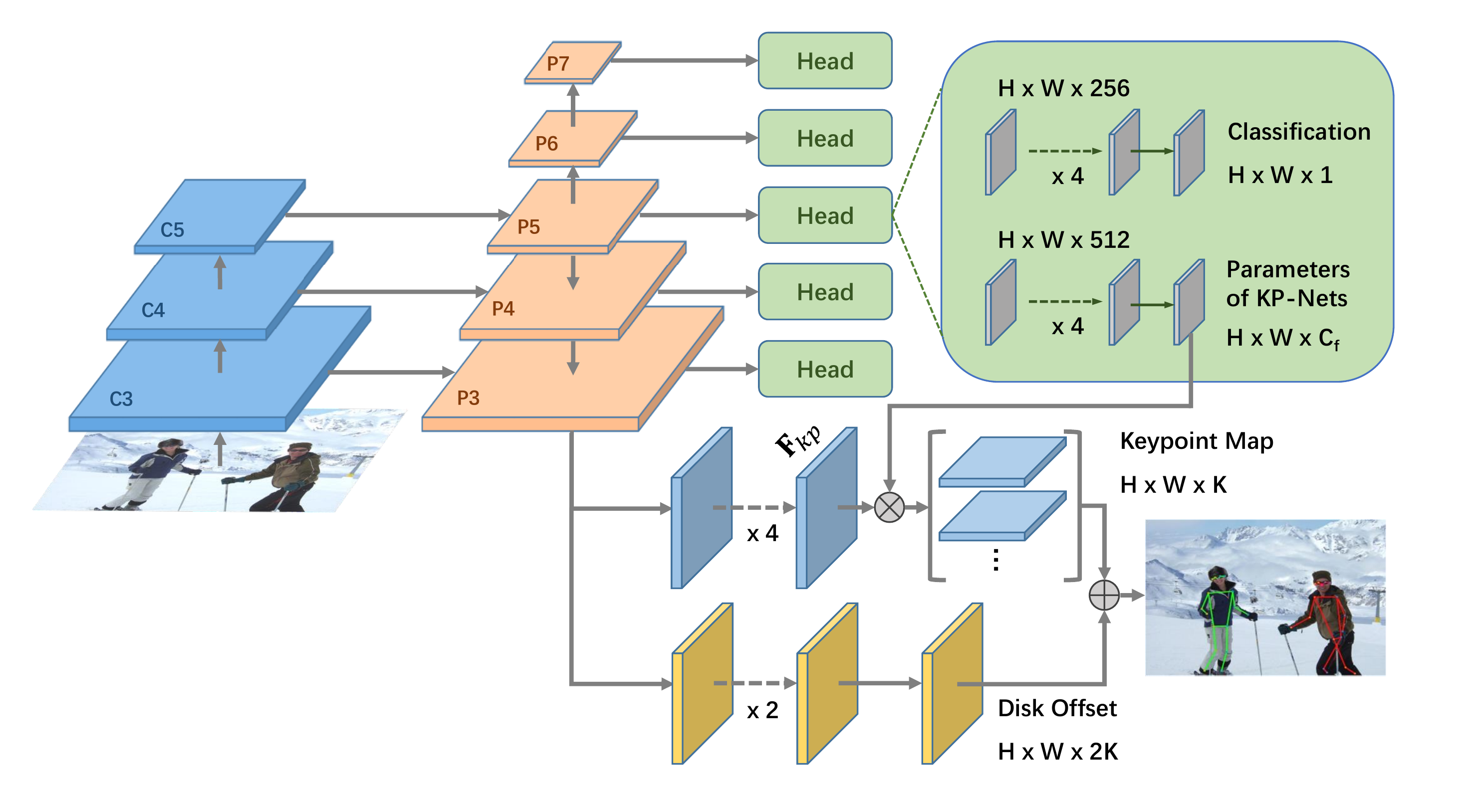}
  \caption{The overall architecture of InsPose. \textit{C}$_{3}$, \textit{C}$_{4}$, \textit{C}$_{5}$ are the feature maps of the backbone network (\textit{e.g.}, ResNet-50). \textit{P}$_{3}$ to \textit{P}$_{7}$ are the FPN feature maps as in \cite{lin2017feature, tian2019fcos}, which are used for final predictions. $H \times W$ denotes the height and width of feature maps, and $K$ is the number of keypoints for each instance. Classification head is used to classify the locations on the feature maps into "person" or "not person".
  The parameters of KP-Nets are dynamically generated with respect to different locations of the input. Then the learned parameters are split and reshaped as the weights and biases of convolutions in each KP-Net which is applied to predict keypoint maps for the particular instance.
  Disk offset branch predicts the instance-agnostic local offset to recover the discretization error caused by down-sampling.}
  \label{fig:CondPose}
\end{figure*}

The overview of our method is illustrated in Figure~\ref{fig:CondPose} and we summarize our main contributions as below.

\begin{itemize}

\item We propose a simple single-stage network for multi-person pose estimation. It is end-to-end trainable and avoids many heuristic designs such as keypoints grouping~\cite{cheng2020higherhrnet}, progressive pose representation~\cite{nie2019single}, delicate feature aggregation~\cite{tian2019directpose}, or pose anchors~\cite{wei2020point}.

\item The proposed model has a dynamic network that consists of instance-aware KP-Nets, which is different from previous methods that use a fixed set of convolutional filters for all the instances.

\item We conduct extensive experiments on the COCO benchmark where our method outperforms state-of-the-art single-stage methods significantly and achieves a better trade-off between accuracy and efficiency compared to state-of-the-art two-stage approaches.

\end{itemize}

\begin{figure*}[t]
  \centering
  \includegraphics[width=\linewidth]{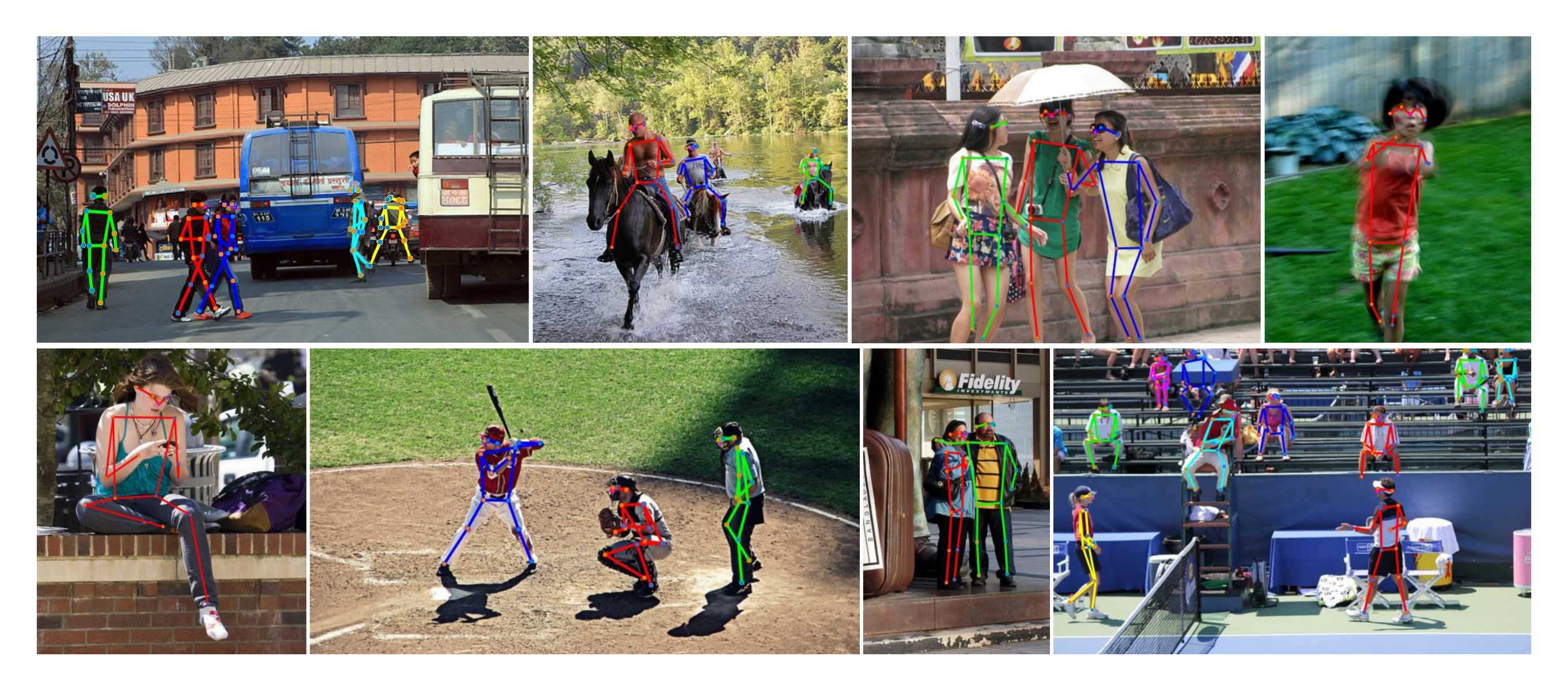}
  \caption{Visualization results of the proposed InsPose on MS-COCO \texttt{val2017}. InsPose can directly detect a wide range of poses, containing viewpoint change, occlusion, motion blur, multiple persons. Magenta, blue, and orange dots represent nose, keypoints of left body, and keypionts of right body, respectively. Note that some small-scale person do not have ground-truth keypoint annotations in the training set of MS-COCO, thus they might be missing when testing. Best viewed in color.}
  \label{fig:CondPoseVisRes}
\end{figure*}

\section{Related Work}
\subsection{Multi-Person Pose Estimation}
With the development of deep neural networks, modern pose estimation methods can be divided into three categories: top-down methods, bottom-up methods, and recent single-stage methods.

\textbf{Top-down Methods.}\quad
Top-down methods decouple the multi-person pose estimation task into two sub-tasks: multi-person detection and single-person pose estimation.
The person detection is first performed to predict a bounding-box for each instance in the image.
Then the instance is cropped from the bounding box for single-person pose estimation.
Instead of cropping the original image, Mask RCNN \cite{he2017mask} proposes to utilize extracted features to improve efficiency.
A cascade pyramid network is proposed in CPN \cite{chen2018cascaded}, aiming to regress difficult keypoints.
Moreover, HRNet \cite{sun2019deep} focuses on learning reliable high-resolution representations and maintains high-resolution representations through the network.
In general, top-down methods have better performance but also have higher computational complexity since they have to repeatedly perform single-person pose estimation for each instance.
Moreover, top-down methods are highly dependent on the performance of the detectors.

\textbf{Bottom-up Methods.}\quad 
On the other hand, bottom-up methods \cite{cao2017realtime}, \cite{insafutdinov2016deepercut}, \cite{newell2016associative}, \cite{cheng2020higherhrnet} first detect all keypoints in instance-agnostic fashion.
Then keypoints are assembled into full-body poses by a grouping process.
Openpose \cite{cao2017realtime} proposes to utilize part affinity fields to establish connections between keypoints of the same instance.
Associative embedding \cite{newell2016associative} produces a detection heatmap and a tagging heatmap for each body joint, and then groups body joints with similar tags into individual people.
Based on the above grouping strategy, HigherHRNet \cite{cheng2020higherhrnet} proposes multi-resolution heatmap aggregation for more accurate human pose estimation.
Bottom-up methods are usually more efficient because of their simpler pipeline of sharing convolutional computation.
However, the grouping post-process is heuristic and involves many tricks which often makes its performance inferior to top-down methods.

\textbf{Single-Stage Methods.}\quad 
Both top-down and bottom-up methods are not end-to-end trainable and have their own limitations.
The single-stage methods \cite{nie2019single}, \cite{zhou2019objects}, \cite{tian2019directpose}, \cite{wei2020point} are proposed recently to avoid the aforementioned difficulties.
These methods predict instance-aware keypoints directly from estimated root locations \cite{nie2019single}, \cite{zhou2019objects} or follow the dense predication strategy \cite{wei2020point}, \cite{tian2019directpose} for estimating a full-body pose from each spatial location.
SPM \cite{nie2019single} proposes a structured pose representation to unify position information of person instances and body joints.
DirectPose \cite{tian2019directpose}
and Point-set anchors \cite{wei2020point}
adopt deformable-like convolutions to refine the initial estimations, mitigating the difficulties of feature mis-alignment.
The performance of DirectPose \cite{tian2019directpose} is poor even with a well-designed feature aggregation mechanism. 
Although SPM and Point-set anchors achieve competitive performance among single-stage methods, they still fall behind the state-of-the-art bottom-up methods.

\subsection{Dynamic Neural Networks}
The dynamic neural network is an emerging research topic in deep learning.
Unlike traditional neural networks which have a static model at the inference stage, dynamic networks can adapt their structures or parameters to different inputs, leading to advantages in terms of adaptiveness, capacity, efficiency, etc.
A typical approach to parameter adaptation is adjusting the weights based on their input during inference.
For example, conditionally parameterized convolution~\cite{yang2019condconv} and dynamic convolutional neural network~\cite{Chen_2020_CVPR} perform soft attention on multiple convolutional kernels to produce an adaptive ensemble of parameters without noticeably increasing the computational cost.
On the other hand, weight prediction~\cite{DBLP:conf/nips/DenilSDRF13} directly generates (a subset of) instance-wise parameters with an independent model at test time. 
For instance, dynamic filter networks~\cite{jia2016dynamic} build a generation network to predict weights for a convolutional layer.
Recently, CondInst~\cite{tian2020conditional} employs the dynamic filters to generate binary masks for each instance and achieves promising performance in the instance segmentation field.

In this work, we leverage this idea to produce instance-aware dynamic convolutional weights to solve the challenging multi-person pose estimation problem, as shown in Figure \ref{fig:CondPoseVisRes}.
In parallel to our work, Mao \textit{et. al} \cite{Mao_2021_CVPR} proposed a fully convolutional multi-person pose estimator.
Our method and \cite{Mao_2021_CVPR} both present a single-stage multi-person pose estimation framework using dynamic instance-aware convolutions, which achieve better accuracy/efficiency trade-off than other state-of-the-art methods.

\section{Approach}
In this section, we first introduce the overall architecture of our framework. Next, we elaborate on the proposed instance-aware dynamic network, which is capable of implementing the multi-person pose estimation task. Then we describe the disk offset prediction branch to reduce the discretization error. Finally, the loss function and the inference pipeline of our model are summarized.

\subsection{Overview}

Let
\begin{math}
  I \in \mathbb{R}^{H \times W \times 3}
\end{math}
be an input image of height $H$ and width $W$, multi-person pose estimation (a.k.a. keypoint detection) aims at estimating human poses $\mathcal{\bar{P}}$ of all the person instances in $I$. The ground-truths are defined as
\begin{equation}
  \mathcal{\bar{P}} = \{P_{i}^{1}, P_{i}^{2}, \ldots, {P_{i}^{K}\}_{i=1}^{N}},
\end{equation}
where $N$ is the number of persons in the image, $K$ is the number of keypoints (\textit{i.e.}, left shoulder, right elbow), and
\begin{math}
  P_{i}^{j} = (x_{i}^{j}, y_{i}^{j})
\end{math}
denotes coordinates of the $j$th keypoint from the $i$th person.

\textbf{Multi-Person Pose Representation.}\quad
In this work, our core idea is that for an image with $N$ instances, $N$ different KP-Nets will be dynamically generated, and each KP-Net will contain the characteristics of the target instance in its filter kernels.
As shown in Figure \ref{fig:CondPosePerInstance}, when the KP-Net for person $i$ is applied to an input, it will produce a keyoint map $M_{i} \in \mathbb{R}^{\frac{H}{s} \times \frac{W}{s} \times K}$, where $s$ is the down-sampling ratio of the map. Then the maximal activation in $j$th channel of ${M_{i}}$ (denoted as ${M_{i}^{j}}$) indicates the location of the $j$th keypoint for person $i$, which can be formulated as
\begin{equation}
  \label{equ:getmax}
  \bar{P}_{i}^{j} = \left( \bar{x}_{i}^{j}, \ \bar{y}_{i}^{j} \right) = \mathop{\arg\max}_{x,\ y} \ M_{i}^{j}
\end{equation}

To recover the discretization error caused by the down-sampling of keypoint maps, we additionally propose a disk offset prediction module. It predicts an instance-agnostic local offset $O \in \mathbb{R}^{\frac{H}{s} \times \frac{W}{s} \times 2K}$ for each category of keypoints. Then we pick the offset vector $(\Delta x, \Delta y)$ corresponding to the above keypoint location $\bar{P}_{i}^{j}$, the final keypoint location on the input image plane can be defined as
\begin{equation}
  \label{equ:offset-shifted}
  \hat{P}_{i}^{j} = \left( (\bar{x}_{i}^{j} + \Delta x) \times s, \  \ (\bar{y}_{i}^{j} + \Delta y)  \times s \right)
\end{equation}

\textbf{Network Architecture.}\quad
We build the proposed InsPose on the popular anchor-free object detector FCOS \cite{tian2019fcos} due to its simplicity and flexibility,
as shown in Figure \ref{fig:CondPose}. We make use of ResNet \cite{he2016deep} or HRNet \cite{sun2019deep} as the backbone network. Then feature pyramid networks (FPN) \cite{lin2017feature} is employed to produce 5 feature pyramid levels \{$P_{3},P_{4},P_{5},P_{6},P_{7}$\}, whose down-sampling ratios are 8, 16, 32, 64, 128, respectively.

As shown in previous works \cite{ren2015faster, he2015spatial}, each location on the FPN's feature maps is considered as a positive sample if it is associated with a ground-truth instance. Otherwise the location is regarded as a negative sample (\emph{i.e.}, background).
Specifically, suppose that the feature maps with a down-sampling ratio of $s$ can be represented as
\begin{math}
  P_{i} \in \mathbb{R}^{\frac{H}{s} \times \frac{W}{s} \times C}
\end{math}.
A location $(x,y)$ on the feature maps $P_{i}$ can be mapped back onto the input image as
\begin{math}
  \left( xs +\lfloor \frac{s}{2} \rfloor, ys +\lfloor \frac{s}{2} \rfloor \right)
\end{math}.
As described in previous works \cite{tian2019fcos, tian2020conditional}, 
if the mapped location falls in the center region of an instance, the location is assigned to be responsible for that instance.
The center region is defined by the box
\begin{math}
  \left( c_{x}-rs, c_{y}-rs, c_{x}+rs, c_{y}+rs \right)
\end{math}, where
\begin{math}
  \left( c_{x}, c_{y} \right)
\end{math}
denotes the pseudo-box (minimum enclosing rectangles of the keypoints) center of the instance.
$r$ denotes the radius of positive sample regions and is a constant being 1.5 as in FCOS \cite{tian2019fcos}.

As shown in Figure \ref{fig:CondPose}, on each feature level of the FPN, some functional heads (in the dashed box) are applied to make instance-related predictions. For example, the classification head predicts the class (\textit{i.e.}, person or not person) of the instance associated with the location.
Note that the parameters of these heads are shared between FPN levels.

\begin{figure}[t]
  \centering
  \includegraphics[width=\linewidth]{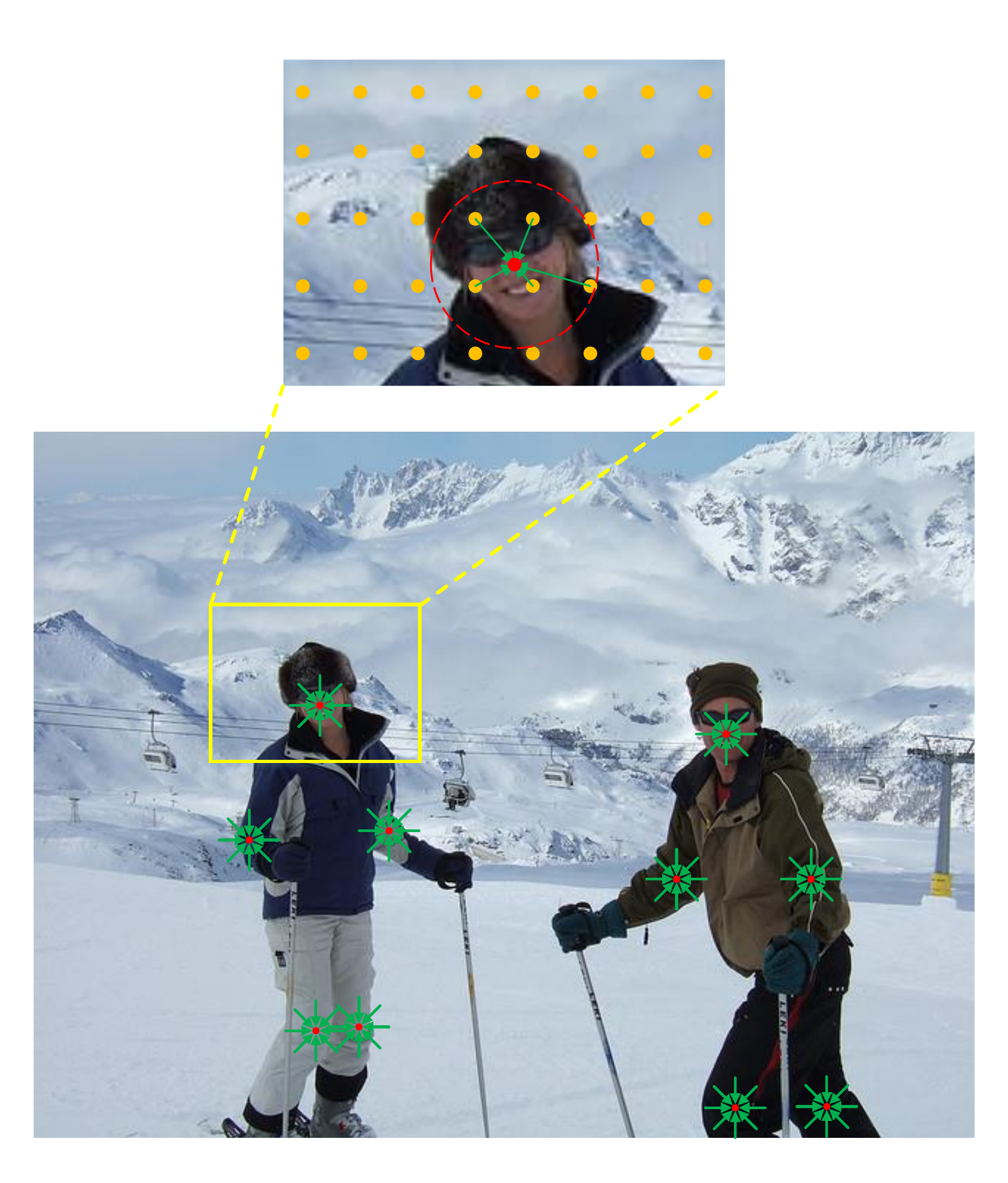}
  \caption{Illustration of disk offset estimation. In the topmost cropped image, orange dots represent feature points while the dotted circle represents the disk region corresponding to the \texttt{nose}. Each position within the disk predicts an offset vector (\textit{i.e.}, green arrow), which is utilized to improve the keypoint localization accuracy. We only draw the keypoints of \texttt{nose, left\_elbow, right\_elbow, left\_knee, right\_knee} for clarity.}
  \label{fig:CondPoseDiskOffset}
\end{figure}

\subsection{Instance-Aware Dynamic Networks}

\textbf{Dynamic KP-Nets.}\quad
To predict the parameters $\bm{\theta}_{x,y}$ of the KP-Net with respect to the location $(x,y)$ of the input feature map, we concatenate all the parameters in each KP-Net (\emph{i.e.}, weights and biases) together as a $C_{f}$-D vector, where $C_{f}$ is the total number of the parameters.
Similar to CondInst~\cite{tian2020conditional}, the KP-Net is a very compact FCN architecture, which has three $1 \times 1$ convolutions, each having 8 channels and using ReLU as the activation function except for the last one.
No normalization layer such as batch normalization \cite{ioffe2015batch} is adopted here. 
The last layer has $K$ (\textit{e.g.}, $K=17$ in COCO \cite{lin2014microsoft} dataset) output channels and generates the final keypoint map for the instance.
In particular, the KP-Net has 313 parameters in total, which consists of conv1  ($(8+2) \times 8 + 8$), conv2 ($8 \times 8 + 8$) and conv3 ($8 \times 17 + 17$).
As mentioned before, the dynamically generated KP-Net contains information about the instance at the location. Thus, it will ideally only fire on the pixels of the instance's keypoints, as shown in Figure \ref{fig:CondPosePerInstance}.

\textbf{Shared Inputs.}\quad
As shown in Figure \ref{fig:CondPose}, there is a keypoint feature branch connected to FPN level $P_{3}$ with a down-sampling ratio of 8, which provides the shared feature map $\textbf{F}_{kp} \in \mathbb{R}^{\frac{H}{s} \times \frac{W}{s} \times C_{kp}} $ (\emph{i.e.}, $s=8$) for the final pose estimation.
It first stacks four $3 \times 3$ convolutions with 128 channels before the last layer.
Then, to reduce the number of parameters in the KP-Net, the last layer of the keypoint feature branch decreases the number of channels from 128 to 8 (\textit{i.e.}, $C_{kp}=8$).
Considering that the relative coordinate map can provide a strong cue for distinguishing different instances, we therefore combine $\textbf{F}_{kp}$ with a relative coordinate map
\begin{math}
  \mathbf{R}_{x,y} \in \mathbb{R}^{\frac{H}{s} \times \frac{W}{s} \times 2}
\end{math},
which are relative coordinates to the location $(x,y)$ (\textit{i.e.}, where the parameters $\bm{\theta}_{x,y}$ are generated).
Afterwards, the combination
\begin{math}
  \mathbf{\tilde{F}}_{x,y} \in \mathbb{R}^{\frac{H}{s} \times \frac{W}{s} \times (C_{kp}+2)}
\end{math}
is fed into the KP-Nets to predict the keypoint map, which is omitted in Figure \ref{fig:CondPose} for simplicity.

\textbf{Optimization.}\quad
The loss function of the instance-aware KP-Nets module can be formulated as:
\begin{equation}
\begin{split}
  & L_{kpf} \left( \left\{ \bm{\theta}_{x,y} \right\} \right) = \\
  & \frac{1}{N_{pos}} \sum\limits_{x,y} \mathbbm{1}_{\{c_{x,y}^{*} > 0\}} L_{ce} \left( FCN \left( \mathbf{\tilde{F}}_{x,y}; \bm{\theta}_{x,y} \right), \mathbf{M}_{x,y}^{*} \right),
\end{split}
\label{eq:loss-kpf}
\end{equation}
where \textit{FCN} denotes the fully convolutional dynamic networks, which consists of the generated instance-aware parameters $\bm{\theta}_{x,y}$ at location $(x,y)$.
The classification label $c_{x,y}^{*}$ of location $(x,y)$ is 1 if the location is associated with an ground-truth person and is 0 for indicating background.
\begin{math}
  \mathbbm{1}_{\{c_{x,y}^{*} > 0\}}
\end{math}
is the indicator function, being 1 if $c_{x,y}^{*} > 0$ and 0 otherwise.
${N_{pos}}$ is the total number of locations where $c_{x,y}^{*} > 0$.
As described before, $\mathbf{\tilde{F}}_{x,y}$ is the combination of the keypoint feature $\mathbf{F}_{kp}$ and the relative coordinate map $\mathbf{R}_{x,y}$.

As mentioned before, the resolution of the predicted keypoint map is $\frac{1}{s}$ of the input image’s resolution. Thus, for each visible ground-truth keypoint of an instance, we compute its low-resolution equivalent $\hat{p}=\left( \lfloor \frac{x^{*}}{s} \rfloor, \lfloor \frac{y^{*}}{s} \rfloor \right)$. The training target is a one-hot $\frac{H}{s} \times \frac{W}{s}$ binary mask where only a \textit{single} pixel at location $\hat{p}$ is labeled as foreground.
And
\begin{math}
  \mathbf{M}_{x,y}^{*} \in \{0,1\}^{\frac{H}{s} \times \frac{W}{s} \times K}
\end{math}
represents the $K$ one-hot masks of the person instance associated with location $(x,y)$. $L_{ce}$ is the softmax cross-entropy loss, which encourages a single keypoint to be detected. These operations are omitted in Equation \eqref{eq:loss-kpf} for clarification.

\begin{table*}
  \caption{Ablation experiments on COCO \texttt{val2017}. ``Disk offset'': the predicted local offset for recovering the discretization error caused by the keypoint map stride. ``Heatmap'': using heatmap to assist training.}
  \label{tab:abl-exp}
  \begin{tabular}{p{60 pt}<{\centering}|p{50 pt}<{\centering}p{50 pt}<{\centering}|p{28 pt}<{\centering}p{28 pt}<{\centering}p{28 pt}<{\centering}p{28 pt}<{\centering}p{28 pt}<{\centering}p{28 pt}<{\centering}p{28 pt}<{\centering}p{28 pt}<{\centering}}
    \toprule
    \ & Disk offset & Heatmap & ${\rm AP}$ & ${\rm AP}_{50}$ & ${\rm AP}_{75}$ & ${\rm AP}_{M}$ & ${\rm AP}_{L}$ & ${\rm AR}$ & ${\rm AR}_{50}$ & ${\rm AR}_{75}$ \\
    \midrule
    Our InsPose & \ & \ & 60.3 & 85.4 & 65.5 & 55.0 & 68.4 & 68.6 & 91.0 & 73.7 \\
    Our InsPose & \checkmark & \ & 62.1 & 86.0 & 67.5 & 57.6 & 69.1 & 70.2 & \textbf{91.4} & 75.3 \\
    Our InsPose & \checkmark & \checkmark & \textbf{63.1} & \textbf{86.2} & \textbf{68.5} & \textbf{58.5} & \textbf{70.1} & \textbf{70.9} & 91.2 & \textbf{76.1} \\
  \bottomrule
\end{tabular}
\end{table*}

\subsection{Disk Offset Estimation}
Suppose that $M_{i} \in \mathbb{R}^{\frac{H}{s} \times \frac{W}{s} \times K}$ is the predicted keypoint map for the $i$th person. 
The location of maximal activation in the $j$th channel of $M_{i}$ indicates the location of the $j$th keypoint of person $i$, 
which is defined as $\bar{P}_{i}^{j} = ( \bar{x}_{i}^{j}, \bar{y}_{i}^{j} )$. 
We can get the final keypoint location by mapping back $\bar{P}_{i}^{j}$ onto input image plane as $((\bar{x}_{i}^{j} + 0.5) \times s, (\bar{y}_{i}^{j} + 0.5) \times s)$.
However, there is a discretization error in this naive position mapping due to the down-sampling operation.
In order to compensate for the discretization error, we predict a instance-agnostic local offset $O \in \mathbb{R}^{\frac{H}{s} \times \frac{W}{s} \times 2K}$ for each category of keypoints.

As shown in Figure \ref{fig:CondPose}, the disk offset branch takes the FPN feature maps $P_{3}$ as input.
Afterwards, three $3 \times 3$ conv layers are applied for the final disk offset prediction.
There are $2K$ output channels in the offset prediction, indicating the displacements in both the horizontal and vertical directions.
Let $p$ be the 2-D position in the feature plane, and
\begin{math}
  \mathcal{D}_{R} = \{ p : || p - P_{i}^{j} || \le R \}
\end{math}
be a disk of radius $R$ centered around $P_{i}^{j}$, where $P_{i}^{j}$ represents the 2-D position of the $j$th keypoint of the $i$th person.
As illustrated in Figure \ref{fig:CondPoseDiskOffset}, for each position $p$ within the disk region $\mathcal{D}_{R}$, the 2-D offset vector which points from the image position $x_{j}$ to keypoint $P_{i}^{k}$ is generated.
During training, we penalize the disk offset regression errors with the $L_{1}$ loss.
The supervision is only performed in the disk region.
During inference, the position of maximum response $(\bar{x}_{i}^{j}, \bar{y}_{i}^{j})$ in the keypoint map is shifted by the corresponding offset vector $(\Delta x, \Delta y)$.
Then the final result can be formulated as
\begin{math}
  ((\bar{x}_{i}^{j}+\Delta x) \times s, (\bar{y}_{i}^{j}+\Delta y) \times s)
\end{math}.
This well-designed module can further improve pose estimation performance, as shown in our experiments.

\subsection{Training and Inference}

\textbf{Training and Loss Fucntion.}\quad
Given the fact that multi-task learning facilitates the training of models, we adopt a multi-person heatmap prediction as an auxiliary task, which is omitted in Figure \ref{fig:CondPose} for simplicity.
The heatmap prediction task takes as input the FPN feature maps $P_{3}$. Afterwards, two $3 \times 3$ convolutions with channel being 256 are applied here. Finally, another $3 \times 3$ convolution layer with the output channel being $K$ is appended for the final heatmap prediction, where $K$ is the number of keypoint categories in the dataset. 
As shown in the following experiments, this joint learning procedure can boost the performance of our model. 
Note that the heatmap-based task is only used for auxiliary supervision during training and \textit{is removed when testing.}

Formally, the overall loss function of InsPose can be formulated as:
\begin{equation}
  L = L_{cls} + L_{kpf} + L_{do} + L_{hm}
\end{equation}
where $L_{cls}$ is the focal loss function \cite{lin2017focal} for the classification branch, $L_{kpf}$ is the loss of instance-aware KP-Nets module as defined in Equation \eqref{eq:loss-kpf}, $L_{do}$ is L1 loss for disk offset regression, $L_{hm}$ is a variant of focal loss \cite{law2018cornernet} for the auxiliary heatmap branch.

\textbf{Inference.}\quad
The inference of InsPose is simple and straightforward. Given an input image, we forward it through the network to obtain the outputs including classification confidence $\bm{p}_{x,y}$ and the generated parameters $\bm{\theta}_{x,y}$ of the KP-Net at location $(x,y)$.
We first use a confidence threshold of 0.1 to filter out predictions with low confidence. Then we select the top $N$ (\text{i.e.}, $N=500$) scoring person instances and their corresponding parameters. Then the N groups of pose estimation results can be obtained through Equations \eqref{equ:getmax} and \eqref{equ:offset-shifted}.
At last, We do non-maximum suppression (NMS) using the minimum enclosing rectangles of estimated keypoints of the instances.
The top 100 instance keypoints in each image are kept for evaluation.

It is worth noting that our framework can conveniently be adapted to simultaneous bounding-box and keypoint detection by attaching a  bounding-box head, which shares the preceding convolutions of the parameter generating branch for KP-Nets.

\section{Experiments}
We evaluate our proposed InsPose on the large-scale COCO benchmark dataset \cite{lin2014microsoft}, which contains more than 250$K$ instances where each person has 17 annotated keypoints. Consistent with previous methods \cite{he2017mask, tian2019directpose, wei2020point}, we use the COCO \texttt{train2017} split (57$K$ images) for training and \texttt{val2017} split (5$K$ images) as validation for our ablation study. The main results are reported on the \texttt{test-dev} split (20$K$ images) for comparison with other methods.
We use the standard COCO metrics including AP, AP$_{50}$, AP$_{75}$, AP$_M$, AP$_L$ to evaluate the performance of pose estimation.

\textbf{Implementation Details:}\quad Unless specified, ablation studies are conducted with ResNet-50 \cite{he2016deep} and FPN \cite{lin2017feature}. 
The models are trained with stochastic gradient descent (SGD) with a minibatch of 16 images.
The initial learning rate is set to 0.01. Weight decay and momentum are set as 0.0001 and 0.9, respectively. Specifically, the model is trained for 36 epochs and the initial learning
rate is divided by 10 at epoch 27 and 33 in ablation experiments.
For the main results on \texttt{test-dev} split, we train the models for 100 epochs and the initial learning rate is divided by 10 at epoch 75 and 90.
ImageNet \cite{deng2009imagenet} pre-trained ResNet \cite{he2016deep} is employed to initialize our backbone networks. We initialize the newly added layers as described in \cite{lin2017focal}. When training, the images are resized to have their shorter sides in $[640, 800]$ and their longer sides less or equal to 1333, and then they are randomly horizontally flipped with the probability of 0.5. No extra data augmentation like rotation \cite{nie2019single} or random cropping \cite{tian2019directpose} is used during training. When testing, the input images are resized to have their shorter sides being 800 and their longer sides less or equal to 1333.

\subsection{Ablation Experiments}

\textbf{Baseline.}\quad
First, we conduct the experiments through employing instance-aware dynamic networks to produce keypoint maps where the maximal activation of each channel is chosen as the keypoint. As shown in Table \ref{tab:abl-exp} (1st row), the naive framework can obtain competitive performance (60.3\% in AP$^{kp}$).
As shown in Figure \ref{fig:CondPose}, we apply the local offset predicted by the disk offset branch to recover the discretization error of keypoint maps. As shown in Table \ref{tab:abl-exp} (2nd row), the performance is improved from 60.3\% to 62.1\% in AP$^{kp}$, which demonstrate the effectiveness of disk offset branch. Therefore, in the sequel, we will use the disk offset branch for all the following experiments.
By jointly learning with a heatmap prediction task, the performance of our model can be further improved from 62.1\% to 63.1\%, as shown in Table \ref{tab:abl-exp} (3rd row). Note that the heatmap prediction is only used during training to provide the multi-task regularization and is pruned during inference. We show some visualization results of InsPose in Figure \ref{fig:CondPoseVisRes}.

\textbf{Architecture of KP-Net.}\quad
In this section, we discuss the architecture of the KP-Nets in InsPose. Our baseline is the KP-Net of three $1 \times 1$ convolutions with 8 channels considering the high capacity of conditional convolutions, which contains a total of 313 parameters.
As shown in Table \ref{tab:abl-depth} (3rd row), it achieves 63.1\% in AP$^{kp}$. Then, we conduct experiments by varying the depth of the KP-Net.
As shown in Table \ref{tab:abl-depth}, the depth being 1 achieves inferior performance as the KP-Net is actually equivalent to a linear mapping, which has overly weak capacity in this case. With the depth increases, the performance is improved gradually due to the higher capacity of the KP-Net. 
However, the performance drops slightly when the depth increases further mainly due to the reason that it is harder to optimize more dynamic parameters.

\textbf{Resolution of Keypoint Maps and Disk Offset Predictions.}\quad
Intuitively, the resolution of the above two predictions is of great importance to the final performance due to pose estimation being a position-sensitive task.
We investigate this in the experiment.
As shown in Figure \ref{fig:CondPose}, the resolution of the origin keypoint score maps and disk offset predictions is same as the resolution of P3 features, which is $\frac{1}{8}$ of the input image resolution.
As shown in Table \ref{tab:abl-res}, our baseline model (2nd row in the table) achieves 63.1\% in AP$^{kp}$.
If both of the two predictions is downsampled to the $\frac{1}{16}$ of the input image resolution (1st row in the table), the performance drops slightly (from 63.1\% to 62.6\%). In paticular, AP$_{M}$ drops remarkably by 1.3\% (from 58.5\% to 57.2\%), which indicating that low-resolution predictions harms the performance of small-scale persons.
Additionally, upsampling both of the predictions by a factor of 2 and 4 (using bilinear interpolation), the performance is degraded by 1.1\% and 16.6, respectively. This is probably due to the relatively low-quality annotations of keypoints, which causes a growing negative impact on the training of the model as the output resolution increases.

\begin{table}
\renewcommand{\arraystretch}{1.2}
  \caption{Ablation experiments on COCO \texttt{val2017} with different depth of KP-Nets. ``depth'': the number of conv. layers in the KP-Nets.}
  \label{tab:abl-depth}
  \begin{tabular}{p{45 pt}<{\centering}p{25 pt}<{\centering}p{25 pt}<{\centering}p{25 pt}<{\centering}p{25 pt}<{\centering}p{25 pt}<{\centering}}
    \toprule
    depth & ${\rm AP}$ & ${\rm AP}_{50}$ & ${\rm AP}_{75}$ & ${\rm AP}_{M}$ & ${\rm AP}_{L}$ \\
    \midrule
    1 & 54.7 & 81.1 & 57.4 & 50.5 & 62.2 \\
    2 & 62.4 & 85.9 & 67.9 & 58.0 & 69.2 \\
    3 & \textbf{63.1} & \textbf{86.2} & \textbf{68.5} & \textbf{58.5} & \textbf{70.1} \\
    4 & 61.7 & 85.9 & 67.2 & 57.1 & 68.7 \\
  \bottomrule
\end{tabular}
\end{table}

\textbf{Combining with Bounding Box Detection.}\quad
By simply attaching a bounding-box regression head, we can simultaneously detect bounding-boxes and keypoints of the instances.
When testing, NMS is applied to the detected bounding-boxes instead of the minimum enclosing rectangles of keypoints. Here we confirm it by the experiment. As shown in Table \ref{tab:abl-box}, our framework can achieve a reasonable person detection performance, which is competitive with the Faster R-CNN detector in Mask R-CNN \cite{he2017mask} (52.5\% \textit{vs.} 50.7\%). Meanwhile, the keypoint performance drops a little due to different positive/negative label assignments between the detection-boxes and pseudo-boxes at training phase.

\begin{table}
\renewcommand{\arraystretch}{1.2}
  \caption{Ablation experiments on COCO \texttt{val2017} with different resolution of keypoint score maps and disk offset predictions. ``res. ratio'': the resolution ratio of the above two predictions to the input image.}
  \label{tab:abl-res}
  \begin{tabular}{p{45 pt}<{\centering}p{25 pt}<{\centering}p{25 pt}<{\centering}p{25 pt}<{\centering}p{25 pt}<{\centering}p{25 pt}<{\centering}}
    \toprule
    res. ratio & ${\rm AP}$ & ${\rm AP}_{50}$ & ${\rm AP}_{75}$ & ${\rm AP}_{M}$ & ${\rm AP}_{L}$ \\
    \midrule
    1/16 & 62.6 & 85.8 & 68.2 & 57.2 & \textbf{70.5} \\
    1/8 & \textbf{63.1} & \textbf{86.2} & \textbf{68.5} & \textbf{58.5} & 70.1 \\
    1/4 & 62.0 & 83.9 & 67.5 & 57.8 & 69.1 \\
    1/2 & 45.4 & 73.6 & 47.0 & 39.1 & 54.7 \\
  \bottomrule
\end{tabular}
\end{table}

\begin{table}
\renewcommand{\arraystretch}{1.2}
  \caption{Ablation experiments on COCO \texttt{val2017} for InsPose with person bounding-box detection. The proposed framework can achieve reasonable person detection results (52.5\% in box AP). As a reference, the Faster R-CNN person detector in Mask R-CNN \cite{he2017mask} achieves 50.7\% in box AP.}
  \label{tab:abl-box}
  \begin{tabular}{p{30 pt}<{\centering}p{22 pt}<{\centering}p{22 pt}<{\centering}p{22 pt}<{\centering}p{22 pt}<{\centering}p{22 pt}<{\centering}p{22 pt}<{\centering}}
    \toprule
    w/ bbox & ${\rm AP}^{bb}$ & ${\rm AP}^{kp}$ & ${\rm AP}_{50}^{kp}$ & ${\rm AP}_{75}^{kp}$ & ${\rm AP}_{M}^{kp}$ & ${\rm AP}_{L}^{kp}$ \\
    \midrule
    \ & - & \textbf{63.1} & \textbf{86.2} & \textbf{68.5} & \textbf{58.5} & 70.1 \\
    \checkmark & 52.5 & 62.5 & \textbf{86.2} & 68.1 & 58.3 & \textbf{70.3} \\
  \bottomrule
\end{tabular}
\end{table}

\begin{table*}[ht]
\renewcommand{\arraystretch}{1.0}
  \caption{Comparison with state-of-the-art methods on MS COCO \texttt{test-dev} dataset. $^\ast$ and $^\dag$ denote multi-scale testing and using additional refinement, respectively.
  We compare the average inference time per image of our methods with other state-of-the-art single-stage and two-stage methods. The time is counted with single-scale testing on a single NVIDIA TITAN X GPU.
  }
  \label{tab:sota}
  \begin{tabular}{p{110 pt}<{\centering}p{110 pt}<{\centering}p{30 pt}<{\centering}p{30 pt}<{\centering}p{30 pt}<{\centering}p{30 pt}<{\centering}p{30 pt}<{\centering}p{30 pt}<{\centering}}
    \toprule
    Method & Backbone & ${\rm AP}$ & ${\rm AP}_{50}$ & ${\rm AP}_{75}$ & ${\rm AP}_{M}$ & ${\rm AP}_{L}$ & Times[s]\\
    \midrule
    \multicolumn{7}{c}{\textbf{Top-down Methods}} \\
    \midrule
    Mask-RCNN \cite{he2017mask} & ResNet-50 & 62.7 & 87.0 & 68.4 & 57.4 & 71.1 & 0.08\\
    CPN \cite{chen2018cascaded} & ResNet-Inception & 72.1 & 91.4 & 80.0 & 68.7 & 77.2 & > 0.65\\
    HRNet \cite{sun2019deep} &  HRNet-w32 &  74.9 &  92.5 &  82.8 &  71.3 &  \textbf{80.9} &  > 2.34\\
    ECSI \cite{su2019multi} &  ResNet-152 & 74.3 & 91.8 & 81.9 & 70.7 &  80.2 & -\\
    RSN \cite{cai2020learning} &  2 $\times$ RSN-50 &  \textbf{75.5} &  \textbf{93.6} &  \textbf{84.0} &  \textbf{73.0} &  79.6 &  > 3.06\\
    \midrule
    \multicolumn{7}{c}{\textbf{Bottom-up Methods}} \\
    \midrule
    CMU-Pose$^\ast\dag$ \cite{cao2017realtime} & 3CM-3PAF (102) & 61.8 & 84.9 & 67.5 & 57.1 & 68.2 &-\\
    AE$^\ast\dag$ \cite{newell2016associative} & Hourglass-4 stacked & 65.5 & 87.1 & 71.4 & 61.3 & 71.5 & 0.52\\
    PersonLab \cite{papandreou2018personlab} & ResNet-101 & 65.5 & 87.1 & 71.4 & 61.3 & 71.5 &-\\
    PifPaf \cite{kreiss2019pifpaf} & ResNet-152 & 66.7 & - & - & 62.4 & 72.9 & 0.26\\
    HigherHRNet \cite{cheng2020higherhrnet} &  HRNet-w32 & 66.4 &  87.5 & 72.8 & 61.2 &  \textbf{74.2} &  0.60\\
    DGCN \cite{qiu2020dgcn} & ResNet-152 &  \textbf{67.4} &  \textbf{88.0} &  \textbf{74.4} &  \textbf{63.6} & 73.0 & > 0.26\\
    \midrule
    \multicolumn{7}{c}{\textbf{Single-stage Methods}} \\
    \midrule
    DirectPose \cite{tian2019directpose} & ResNet-50 & 62.2 & 86.4 & 68.2 & 56.7 & 69.8 &-\\
    DirectPose \cite{tian2019directpose} & ResNet-101 & 63.3 & 86.7 & 69.4 & 57.8 & 71.2 &-\\
    CenterNet \cite{zhou2019objects} & Hourglass-104 & 63.0 & 86.8 & 69.6 & 58.9 & 70.4 & 0.16\\
    SPM$^\ast\dag$ \cite{nie2019single} & Hourglass-8 stacked & 66.9 & 88.5 & 72.9 & 62.6 & 73.1 &-\\
    Point-Set Anchors$^\ast$ \cite{wei2020point} & HRNet-w48 & 68.7 & 89.9 & 76.3 & 64.8 & 75.3 & 0.27\\
    Our InsPose & ResNet-50 & 65.4 & 88.9 & 71.7 & 60.2 & 72.7 & 0.08\\
    Our InsPose$^\ast$ & ResNet-50 & 67.1 & 89.7 & 74.0 & 62.6 & 73.8 &-\\
    Our InsPose & ResNet-101 & 66.3 & 89.2 & 73.0 & 61.2 & 73.9 & 0.10\\
    Our InsPose$^\ast$ & ResNet-101 & 67.8 & 90.3 & 74.9 & 64.3 & 73.4 &-\\
    Our InsPose & HRNet-w32 & \textbf{69.3} & \textbf{90.3} & \textbf{76.0} & \textbf{64.8} & \textbf{76.1} &  0.15\\
    Our InsPose$^\ast$ &  HRNet-w32 &  \textbf{71.0} &  \textbf{91.3} &  \textbf{78.0} &  \textbf{67.5} & \textbf{76.5} & -\\
  \bottomrule
\end{tabular}
\end{table*}

\subsection{Comparisons with State-of-the-art Methods}
In this section, we evaluate the proposed single-stage multi-person pose estimator on MS-COCO \texttt{test-dev} and compare it with other state-of-the-art methods. 
As shown in Table \ref{tab:sota},  without any bells and whistles (\textit{e.g.}, multi-scale and flipping testing, the refining in \cite{cao2017realtime, newell2016associative}, and any other tricks), the proposed framework achieves 65.4\% and 66.3\% in AP$^{kp}$ on COCO \texttt{test-dev} split, with ResNet-50 and ResNet-101 as the backbone, respectively. With multi-scale testing, our framework can reach 71.0\% in AP$^{kp}$ with HRNet-w32.

\textbf{Compared to Single-stage Methods:}\quad Our method significantly outperforms existing single-stage methods, such as DirectPose \cite{tian2019directpose}, CenterNet \cite{zhou2019objects}, SPM \cite{nie2019single} and Point-Set Anchors \cite{wei2020point}. The performance of our method is +3.2\% and +3.0\% higher as compared to DirectPose \cite{tian2019directpose} with ResNet-50 and ResNet-101, respectively. InsPose with HRNet-w32 even outperforms the latest single-stage pose estimator Point-Set Anchors \cite{wei2020point} with HRNet-w48, 71.0\% \textit{vs.} 68.7\% in AP$^{kp}$. Note that our model is anchor-free and Point-Set Anchors need 27 carefully chosen pose anchors per location.

\textbf{Compared to Bottom-up Methods:}\quad Our method outperforms the state-of-the-art bottom-up methods, such as CMU-Pose \cite{cao2017realtime}, AE \cite{newell2016associative}, PersonLab \cite{papandreou2018personlab}, PifPaf \cite{kreiss2019pifpaf}, HigherHRNet \cite{cheng2020higherhrnet} and the latest DGCN \cite{qiu2020dgcn}. With single-scale testing, it achieves significant improvement over HigherHRNet, 69.3\% \textit{vs.} 66.4\% in AP$^{kp}$, using the same backbone HRNet-w32. Besides that, our performance with HRNet-w32 is higher than DGCN \cite{qiu2020dgcn} with ResNet-152, 69.3\% \textit{vs.} 67.4\% in AP$^{kp}$.

\textbf{Compared to Top-down Methods:}\quad With the same backbone ResNet-50, the proposed method outperforms previous strong baseline Mask R-CNN (65.4\% \textit{vs.} 62.7\% in AP$^{kp}$). Our model is still behind other top-down methods, since they mitigate the scale variance of persons by cropping and resizing the detected person into the same scale. As noted before, two-stage methods take an additional person detector and employ a singe person pose estimator to each cropped image, resulting in high computation overhead. In contrast, our method is much simpler and faster since we eliminate the redundant person detector.

\textbf{Inference Time Comparison:}\quad
We measure the inference time of all variants of our method and other methods on the same hardware if possible.
As shown in Table \ref{tab:sota},
InsPose with HRNet-w32 can infer faster than Point-Set Anchors \cite{wei2020point} (0.15s \textit{vs.} 0.27s per image).
And it also runs significantly faster than state-of-the-art top-down methods, such as HRNet \cite{sun2019deep} (>2.34s per image) and RSN \cite{cai2020learning} (>3.06s per image).
Note that the time of additional person detector used in HRNet and RSN is not included here.
Moreover, we notice that the grouping post-processing in the bottom-up method HigherHRNet \cite{cheng2020higherhrnet} (0.60s per image) is time-consuming and results in inferior speed compared to InsPose.
Overall, InsPose makes a good trade-off between the accuracy and efficiency, benefiting from its simple single-stage framework.

\section{Conclusion}
In this paper, we present a single-stage multi-person pose estimation method, termed InsPose.
It directly maps a raw input image to the desired instance-aware poses, get rid of the need for the grouping post-processing in bottom-up methods or the bounding-box detection in top-down approaches.
Specifically, we propose an instance-aware module to adaptively adjust (part of) the network parameters for each instance.
Our method can significantly increase the capacity and adaptive-ability of the network for recognizing various poses, while maintaining a compact end-to-end trainable pipeline.
We also propose a disk offset branch to recover the discretization error due to down-sampling, boosting the keypoint detection performance further.
We conduct extensive experiments on the MS-COCO dataset where our method achieves significant improvement over existing single-stage methods and performs comparably with state-of-the-art two-stage methods in terms of accuracy and efficiency.


\bibliographystyle{ACM-Reference-Format}
\bibliography{sample-base}

%
%
%

\end{document}